\documentclass[sigconf]{acmart}
\renewcommand\footnotetextcopyrightpermission[1]{} 
\usepackage{booktabs} 

\usepackage{rotating}
\usepackage{multirow}
\usepackage{natbib}
\usepackage{graphicx}
\usepackage{amsmath}

\usepackage[caption=false,font=footnotesize,labelfont=sf,textfont=sf]{subfig}
\def\subfigure{\subfloat}

\usepackage{url}

\newtheorem{mydef}{Definition}

\acmConference{ }{December 2017}{ } 

\begin{document}
\title{Tracking the Diffusion of Named Entities}
\author{Matthew Rowe}
\affiliation{
	\institution{Experian}
	\city{London, UK}
}
\author{Leon Derczynski}
\affiliation{
	\institution{University of Sheffield}
	\city{Sheffield, UK}
}

\begin{abstract}
Existing studies of how information diffuses across social networks have thus far concentrated on analysing and recovering the spread of deterministic innovations such as URLs, hashtags, and group membership.
However investigating how mentions of real-world entities appear and spread has yet to be explored, largely due to the computationally intractable nature of performing large-scale entity extraction.
In this paper we present, to the best of our knowledge, one of the first pieces of work to closely examine the diffusion of named entities on social media, using Reddit as our case study platform.
We first investigate how named entities can be accurately recognised and extracted from discussion posts. 
We then use these extracted entities to study the patterns of entity cascades and how the probability of a user adopting an entity (i.e. mentioning it) is associated with exposures to the entity.
We put these pieces together by presenting a parallelised diffusion model that can forecast the probability of entity adoption, finding that the influence of adoption between users can be characterised by their prior interactions -- as opposed to whether the users propagated entity-adoptions beforehand.
Our findings have important implications for researchers studying influence and language, and for community analysts who wish to understand entity-level influence dynamics.
\end{abstract}


\maketitle

\interfootnotelinepenalty=8000

\section{Introduction}\label{sec:introduction}
Understanding who influences whom and under what conditions forms a core component of information diffusion studies.
Recovering the so-called \emph{diffusion process} allows us to understand how messages are passed between users, and what contributes to their flow.
In turn, this allows for simulation and predictive models to be engineered that forecast the expected spread of information, allowing \emph{spread potential} to be maximised or minimised.

To date, studies of information diffusion on social media and in social networks have concentrated on tracking URLs (e.g. through retweets), link-creation between blogs, hashtags being adopted over time, and group membership adoption; and with different diffusion mechanisms under the microscope (e.g. social contagion, homophily, social reinforcement, rumour spread, structural equivalence, etc.).
Despite the rise in such studies, and in tandem the proliferation of data over which studies can be performed, as yet, and to the best of our knowledge, no work has tracked the spread of entity mentions over time -- a `\emph{named entity}' here being a proper noun representing a person, place, organisation, or something similar.
Understanding \emph{how} named entities diffuse through social networks and being able to \emph{predict} their adoption would provide valuable insights into how topics emerge and spread.

The aim of this paper is to understand how named entities diffuse through social media based discourse, using the online community platform Reddit as the focus of our work.
However, in order to study named entities and how they diffuse, we must answer the following three research questions: \textbf{RQ1:} How can we accurately detect named entities in social media based discourse, given its myriad formats, often informal vernacular, and inherent noise (e.g. misspellings, abbreviations, etc.)? \textbf{RQ2:} What process governs the spread of entities? And how does such spread occur? \textbf{RQ3:} How can we predict the spread of named entities and who will begin talking about them?	

We explored the above questions by devising an approach to recognise entities found in community message board posts -- using the popular site Reddit~\cite{duggan20136} as our study platform.
Using the recognised named entities we then carried out a study of how such entities were adopted over time, how they spread, and created an approach to (accurately) predict the adoption of named entities by users based upon the computation of influence probabilities (e.g. achieving $ROC$ value of $0.755$ in one instance).
The contributions that we make in this paper are as follows:

\begin{enumerate}
	\item A method to recognise and extract named entities for Reddit based upon structured prediction and Brown clustering, together with an evaluation of this method.
	\item A study of how entities spread and are adopted following exposures, using an approach based upon graph isomorphism to track patterns of entity diffusion.
	\item A parallelised general threshold diffusion model, based on the work of Goyal et al.~\cite{goyal2010learning}, that incorporates different entity-adoption constructs (entity propagation, influence of interactions, community homophily) when calculating adoption probabilities -- this is accompanied by a comparative empirical evaluation of the different constructs when forecasting entity adoption within the diffusion process.
\end{enumerate}

We have structured the paper as follows: in the following section we cover related work within the areas of named entity recognition and information diffusion -- paying particular attention in the latter's case to existing works that are \emph{similar} to entity diffusion.
In Section~\ref{sec:datasets_ner} we explain the preparation of the Reddit dataset for our experiments -- including down-sampling of 100 subreddits -- and the adapted named entity recognition (NER) methodology that we employed.
Section~\ref{sec:diffusion} presents findings from our analysis of entity cascades (i.e. their shapes and forms) and how exposure frequency and entity-adoption probability are associated.
This section also describes our implementation of the parallelised general threshold diffusion model and experiments that assess the efficacy of various influence constructs in the entity-diffusion process.
Section~\ref{sec:discussions} discusses the findings that we have drawn from this work and plans for future work, and section~\ref{sec:conclusions} finishes the paper with our conclusions.

%
%
%
%


\section{Related work}
\label{sec:rw}
In this work we investigate how entities diffuse over time through the online community platform Reddit.
Diffusion of information is a well studied topic, and is of particular interest today given the myriad ways in which Web users consume information and are thus influenced by what they read, and with whom they interact with, online.
We first review state of the art approaches for recognising named entities, before then describing existing works that have studied information diffusion.

\subsection{Named Entity Recognition}
The goal of Named entity recognition (NER) is to extract mentions of certain types of entities, like organisations, locations or person names.
Generally, NER systems can be structured in terms of representation, induction, dependency modelling and integration of real-world knowledge~\cite{nadeau2007survey,ratinov2009design}.

While initially conducted over newswire~\cite{tjong2003introduction}, older tools tend not to perform so well on modern text types, such as tweets and other short social media text~\cite{derczynski2015analysis}.
Simultaneously, the value of non-newswire data has increased: social media now provides us with a sample of all human discourse, 
in digital format. This opens areas of investigation such as {\em computational social science}, examining e.g. demographics~\cite{hovy2015user}, personality~\cite{plank2015personality} and income~\cite{preoctiuc2015studying}.

NER for social media content is however difficult, leading to much work, including general approaches~\cite{ritter2011named}, topic-specific approaches~\cite{liu2011recognizing}, adapting from known genres~\cite{plank2014adapting}; these are driven by and evaluated in multiple recent shared tasks~\cite{rowe2015microposts2015,baldwin2015shared}.
The task is generally cast as a domain adaptation problem from newswire data, integrating the two kinds of data for training~\cite{cherryunreasonable} or including a lexical normalisation step~\cite{han2011lexical} to shift text to territory more familiar to existing models and methods.
Major challenges are that NEs mentioned in tweets change over time~\cite{fromreide2014crowdsourcing}, and that diversity of context makes NER more difficult~\cite{derczynski2015analysis}. 
This paper addresses NER without using large amounts of labelled in-domain data, in order to track entity propagation at scale.

\subsection{Information Diffusion}
Studies of information diffusion have largely concentrated on \emph{deterministic} signals of diffusion such as tracking URLs, hashtags, quotes~\cite{suen2013nifty}, and adoption behaviour (e.g. group signups); however to the best of our knowledge such studies have yet to focus on how entities diffuse.
We now focus on key pieces of work that are closely-aligned to the study of entity-diffusion in the context of social networks -- should the reader wish to know more about information diffusion models, and in greater detail, then please refer to Guile et al.'s~\cite{guille2013information} comprehensive survey of such models.

The study of information adoption and sharing was undertaken by Bakshy et al.~\cite{bakshy2012role} who conducted a large-scale randomised controlled trial to examine the effects of \emph{information exposure} on information diffusion, using the Facebook platform.
The authors were able to assign Facebook users \emph{randomly} with a $\frac{1}{3}$ probability to a \emph{feed} group, and the remainder to a \emph{no feed} group and then \emph{hide} information (i.e. status posts) posted within the latter's group.
Bokshy et al. found that users who were \emph{exposed} to information (i.e. those in the feed group) from their friends are more likely to share it on -- implying that such exposure has an influential effect.

The closest work to the study of \emph{entity diffusion} can be found in studies of hashtag diffusion.
For instance, Romero et al.~\cite{romero2011differences} studied the spread of the top-500 hashtags posted in a sample of $>3$B tweets collected over a six-month period, finding that users were most likely to \emph{adopt} a hashtag (i.e. mention/cite it in their Tweet) after receiving $4$ exposures from their friends.
The authors found marked differences in the adoption of hashtags based on their topics, something which -- as we will show below -- is not present in entity diffusion.
More recent work by Yang et al.~\cite{yang2012we} studied both the role of hashtag content and the role of the hashtag in a community, finding that both factors are associated with hashtag adoption.
Our work differs from~\cite{yang2012we} by studying the adoption of entities based on pairwise interactions between users -- i.e. how one user influences another to adopt an entity -- as opposed to the content properties of the entity.

The different modalities of diffusion signals encompass the adoption of behaviour by users from previous adopters, for instance the work of Goyal et al.~\cite{goyal2010learning} tracked the diffusion of \emph{actions} on Flickr, where actions were defined as users joining a group (i.e. a photography-topical group).
A general threshold model was proposed that determines the probability of influence between two arbitrary users based on the relative frequency of action propagations observed before, divided by the absolute number of actions of the user responsible for the propagation.
The authors found that computing the average time of influence between two users led to more accurate computation of influence probabilities.
In this paper, we use the general threshold framework from~\cite{goyal2010learning}, but extend it into the entity-mention setting, hence: we track the \emph{mention} of an entity by a user over time and calculate the probability of influence that an \emph{adopter}'s neighbours have had upon him.
Furthermore, we also extend this framework to test for two additional constructs: (i) influence of interactions before adoption (i.e. did the degree to which an individual communicated with a previous entity adopter influence their own adoption?), and (ii) community homophily (i.e. does the similarity between users' interests -- based upon similarity in subreddit posting -- have an effect on adoption of an entity?).

Prior work on Reddit has examined the site's evolution since launch, seeing it evolve from a bulletin-like page to a large community site with many segragated and unique sub-communities that reinforce a general perception of the overall community~\cite{singer2014evolution}.
This observation supports the use of Reddit as a study venue for information diffusion, finding that communities are large, well-defined, and cohesive.
Later work covers the mapping of popular content~\cite{weninger2015random} and of network structure~\cite{olson2015navigating}, though not the diffusion of information through those networks.

Fang et al.~\cite{fang2013predicting} predicted adoption probabilities in social networks by controlling for potential confounding, unobservable variables -- proposing a modification of expectation-maximisation to induce a Naive bayes predictive model.
The authors found that social influence alone is insufficient to recover the diffusion process, and thus external factors -- that are latent -- must be countered for within any predictive model -- this was in the context of predicting the adoption of social ties.
The adoption of information within a social network and its propagation was studied by Huang et al.~\cite{huang2014temporal} by considering the role of temporal dynamics.
The authors found that the probability of diffusion between users (\emph{retweets} on Chinese microblogging platform Sina Weibo) reduces as a function of time from the last interaction between the users, thereby suggesting that \emph{temporal dynamics} have a strong effect in diffusion.
We build time \emph{explicitly} into our adaptation of Goyal et al.'s~\cite{goyal2010learning} general threshold diffusion model -- by comparing static and discrete time versions of adoption probabilities.

%

\section{Datasets preparation and NER}
\label{sec:datasets_ner} 
To study entity diffusion at a \emph{large-scale} we used the entire dump\footnote{\url{https://archive.org/details/2015_reddit_comments_corpus}} of Reddit from its inception through to July 2015 -- this provided a dataset of $140$Gb of data compressed containing $\sim1.7$B posts (i.e. original thread starter posts and comments).
We also required datasets from which we could \emph{model} and \emph{train} our named entity recogniser -- and also assess its performance -- and used the following:
(i) \emph{CoNLL 2003 data}, a corpus of newswire texts, annotated for named entity chunks and types -- this describes where entity mentions are in the text, including locations, organisations, and person mentions; 
(ii) \emph{Twitter data (unannotated)}, comprised of a large corpus of English tweets taken from an archive of the garden hose feed, and; 
(iii) \emph{Twitter data (annotated)}, comprised of datasets annotated with named entities -- one from Ritter's 2011 EMNLP paper~\cite{ritter2011named} and a second from W-NUT 2015 shared task~\cite{baldwin2015shared}.

For the experiment, we needed to convert the full $140$Gb of compressed Reddit posts into a set of interlinked and time-ordered conversations and the entities mentioned in each of them.
This provides a number of sub-challenges: sampling of the Reddit data, creating a linked series of conversations, and picking out entity mentions in this text type.
Given the lack of prior work on Reddit text, there are no annotated datasets available, so supervised in-domain work is not directly possible.

\begin{figure}
\centering
\includegraphics[width=0.7\columnwidth]{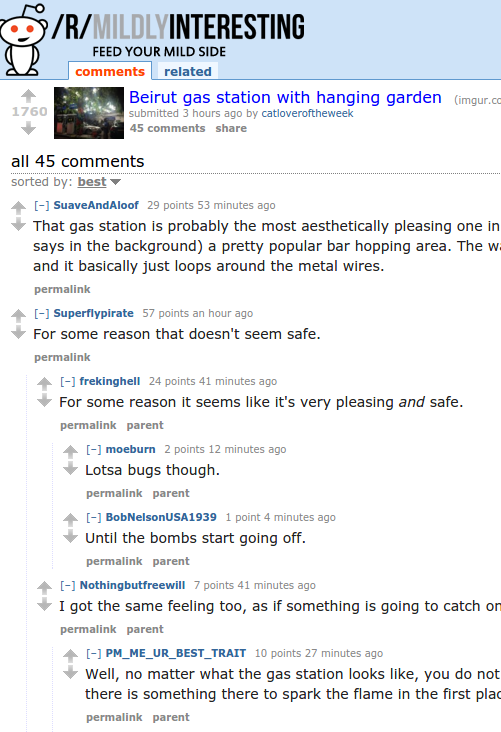}
\caption{Example Reddit post. Note topic at the top, then comments, with conversations following in a tree.}
\label{fig:reddit-example}
\end{figure}

Reddit is roughly similar to a forum, where top-level divisions are made by topic.
Within each topic, or {\em subreddit}, there are posts, which begin with either a short piece of text or a link to an external resource -- typically an image, video, or interesting article.
Users then may publish comments for each post, and reply to each others' comments.
This leads to a threaded discussion, centred on a particular topic, with a hierarchical comment structure (see Figure~\ref{fig:reddit-example}).
The Reddit dataset~\cite{reddit-data} comprises a sequence of comments, with one JSON record for each one, ordered temporally.

\subsection{Subreddit extraction}

The Reddit dataset is large, and had to be pared down for initial analysis.
The data is segmented by community, meaning that the scope of the sample is determined by the selection of {\em subreddits}.
This contrasts, for example, with Twitter, where reducing the sample is performed by reducing the sampling of posts~\cite{kergl2014endogenesis}, thus leading to broken conversation threads and so on.
We chose to examine one hundred entire individual subreddit communities.
The subreddits were chosen from a list of top subreddits (\url{http://redditlist.com/}) which ranks communities based on levels of activity, numbers of subscribers, and rates of growth.
The list of chosen communities can be found in the github repository of this work.\footnote{\url{https://github.com/mrowebot/NER-Diff-Paper}}


\subsection{NER for Reddit}

In this diffusion analysis, we model micro-topics in conversation as entity mentions.
This allows tracking of topics at maximally fine granularity, looking at each user's interests at a low level, as opposed to monitoring broader topics such as ``consumer electronics", ``politics" and so on.
In fact, these broader topics are already explicitly annotated by means of the subreddit topics.

Entity mentions are extracted through named entity recognition.
Generally, this task aims to detect the boundaries of certain kinds of entities within a certain piece of text.
In this instance, we tokenise text, splitting it into sentences using the Punkt tokeniser~\cite{kiss2006unsupervised}, and subsequently word-sized chunks, using the {\tt twokenize} tool with some adaptations~\cite{o2010tweetmotif}.
This tool performs Penn Treebank-style tokenisation, a common standard, with some specific adaptations to enable it to handle the noise present in user-generated text.
After this, we take a structured prediction approach to deciding which tokens in each sentence are part of an entity, and possibly the type of the entity.
Finally, we concatenate entity tokens, and use these to build a list of entity mentions in any given input text.
For example, given the input comment from the source JSON:

\begin{quote}
\emph{``body": ``There are still good fighters on this card. Conor McGregor is there and so is Gunnar Nelson."}
\end{quote}

The following output entities should be collected:

\begin{quote}
\emph{``entity\_texts": [``Conor McGregor", ``Gunnar Nelson"]}
\end{quote}

We present named entity recognition here, adapted and applied to Reddit posts and conversations, a text type for which to our knowledge there have been no prior NER efforts.
Notably, we experiment with techniques previously demonstrated to be successful on other user-generated content and find them lacking.

Machine-learning based NER systems are typically supervised -- they use training data from which features are extracted to form training instances.
However, in general, language processing systems can be hard to transfer between text types; for example, NER systems for newswire might reach about 89\% F1 on news articles, but only around 40\% on tweets (a form of user-generated content), as found in~\cite{derczynski2015analysis}.
One approach to overcome this performance drop when changing text type is to train over a blend of text types.
For example, Ritter et al.~\cite{ritter2011named} used both chat 
and newswire data when developing a part-of-speech tagger for tweets, as well as an unsupervised language model from the target text type.
This led to strong performance improvements.
We follow a similar approach, using a mixture data from both newswire and tweets.
The newswire data is drawn from the CoNLL-2003 evaluation task set~\cite{tjong2003introduction}; the Twitter data is from Ritter et al.'s early experiments and also the W-NUT 2015 shared task~\cite{ritter2011named,baldwin2015shared}.

We start using structured predicting in the form of a Conditional Random Fields (CRF) model to label whole sentences at a time.
For features, we use a fairly classical set, and add some unsupervised word representations to this.
Our base features are: lower-case word; word prefix and suffix (2- and 3-character); previous and next word; flags set if the word is uppercase, titlecase, or digits; these flags for the previous and next words; the next and prior bigrams.

In addition, we induce Brown clusters~\cite{brown1992class} and use these as word representations~\cite{turian2009preliminary}.
Brown clustering is a form of hierarchical agglomerative hard clustering, using average mutual information as a global objective function.
It takes as input a corpus, in the form of a sequence of words, and in its generalised form~\cite{derczynski2016generalised}, a single hyperparameter: the size of its active set $a$.
This active set is filled with the $a$ most-frequent classes drawn from all word classes $C$, with one word per class at initialisation.

The mutual information of two classes, $C_i,C_j\in C$, denoted $MI(C_i,C_j)$, is:

\begin{equation}
MI(C_i,C_j)=
p(\left<C_i,C_j\right>)\ \log_2{\frac{p(\left<C_i,C_j\right>)}{p(\left<C_i,*\right>)\ p(\left<*,C_j\right>)}}
\end{equation}

The average mutual information of $C$, denoted $AMI(C)$, is the sum of mutual information of all cluster pairs in $C$:
\begin{equation}
AMI(C) = \sum_{C_i,C_j\in C}{MI(C_i,C_j)}
\end{equation}

Brown clustering works by greedily merging the pair of classes within the active set that causes the least loss to average mutual information, until all classes are merged.
The result is a sequence of binary merges, describing the set membership of each word type in the corpus as the merges progress.
For each single leaf class, the path to a destination cluster can be described as a bitstring, which details the sequence of binary merges taken.
The zero-length bitstring describes the top of the hierarchy, where there is one class.

These bitstrings are typically converted to features by \emph{shearing}~\cite{derczynski2016generalised}.
This involves only examining the first $n$ bits of a bitstring.
However, shearing does not maximise the information preserved in the representation -- sub-clusterings at many levels are lost.
Therefore we take the cluster identifier at every level, tracing the provenance of a terminal word cluster all the way to the root cluster.
For example, given the bitstring {\tt 1100101}, the following text features are generated: {\tt 1}, {\tt 11}, {\tt 110}, {\tt 1100}, {\tt 11001}, {\tt 110010}, {\tt 1100101}.
If the typical bit depths~\cite{ratinov2009design} of 4, 6, 10 and 20 were chosen, only the following features would be generated: {\tt 1100}, {\tt 110010}, {\tt 1100101}.
As a result of taking all directly-relevant features in the merge list, the lossy nature of shearing-based feature extraction from Brown clusters is avoided.
Feature extraction, training, classification and JSON annotation are all performed using an entity recognition toolkit~\cite{derczynski2015usfd}.

\begin{table}
\small
\centering
\caption{Reddit NER, varying training text type and Brown cluster source. Best per scenario is starred; best overall, bold.}
\begin{tabular}{lrrrr}
\hline
{\bf Brown cluster source} & {\bf Precision} & {\bf Recall} & {\bf F1} & {\bf F2} \\
\hline
\multicolumn{5}{c}{\emph{Baseline}} \\
\hline
Stanford (3-class builtin)   & {\bf 87.88} & 38.93 & 53.96 & 47.81 \\
\hline
\multicolumn{5}{c}{\emph{Newswire training data}} \\
\hline
RCV Newswire          & 63.57 & 59.73 & 61.59 & 60.96 \\
Tweets                & 62.75 & 64.43 & 63.58 & 63.86 \\
Blended tweets/news   &*68.42 & 61.07 & 64.54 & 63.34 \\
Reddit                & 66.32 & {\bf 67.11} & {\bf 66.71} & {\bf 66.84} \\
Stanford baseline     & 63.97 & 58.39 & 61.05 & 60.14 \\
\hline
\multicolumn{5}{c}{\emph{Twitter training data}} \\
\hline
RCV Newswire          &*73.02 & 30.87 & 43.39 & 38.22 \\
Tweets                & 70.37 & 38.26 & 49.57 & 45.12 \\
Blended tweets/news   & 65.28 & 31.54 & 42.53 & 38.10 \\
Reddit                & 76.34 &*47.65 &*58.68 &*54.47 \\
Stanford baseline     & 65.22 & 30.20 & 41.28 & 36.78 \\
\hline
\multicolumn{5}{c}{\emph{Blended training data, 50:50}} \\
\hline
RCV Newswire          & 66.67 & 42.96 & 52.25 & 48.74 \\
Tweets                & 66.10 & 52.35 & 58.43 & 56.25 \\
Blended tweets/news   & 68.69 & 45.64 & 54.84 & 51.39 \\
Reddit                &*70.08 &*59.73 &*64.49 &*62.82 \\
Stanford baseline     & 67.77 & 55.03 & 60.74 & 36.78 \\
\hline
\end{tabular}
\label{tab:brown-tuning}
\end{table}

\subsection{Tuning entity recognition}

Entity recognition needs to be tuned to fit Reddit data well.
Parameters -- in terms of training data composition, feature extraction, and objective function -- should reflect the nature of the task.

For this task, recall is preferable to precision.
Over the large dataset, spurious entities (i.e. false positives) are likely to be seen rarely.
Mis-recognised entity names tend not to be distributed in a few high-frequency clumps, but rather as many different terms, each with a lower frequency.
This suggests that there will be great variation in their surface forms, leaving them in the long tail of entities discovered~\cite{leginus2015enhanced}.
As our diffusion analysis concerns the more frequent entity lexicalisations, these infrequent spurious mis-recognitions are less likely to have an impact.
Indeed, this was borne out in our analysis of entities extracted, with no individual  false-positive, spurious entity surface forms occurring often enough to reach our lists of entities selection for diffusion analysis.
Conversely, recall expresses how broadly and comprehensively the extraction is performing, and is important to tracking a range of entities.\footnote{NB. It has often been more challenging to achieve high recall in social media texts than high precision~\cite{ritter2011named,derczynski2015analysis}.}
That is to say, the problem addressed is more tolerant to low precision in input data than low recall.
We can therefore better evaluate our systems using an adjusted $F_\beta$ score.

\begin{equation}
F_\beta = (1+\beta^2)\frac{PR}{(\beta^2 P) + R} 
\end{equation}

When $\beta=1$, precision and recall are balanced in a harmonic mean, e.g. F1-score.
That is, false positives and false negatives impact results equally.
Given precision $P$ and recall $R$, typically an F-score is drawn from $F_\beta$ with $\beta=1$.
To score away from false negatives, i.e. missed entity mentions, we set $\beta=2$.

Our approach here is to tune an entity recogniser with reference to a dataset that matches the target text type.
We draw this development set from Reddit posts, using comments encountered during our work that appear to have missing or spurious annotations.
These are then isolated, tokenised, and manually annotated.
In total we annotated $3 708$ tokens of Reddit data, having $149$ entity chunks.
This comprised our development set which was used to tune parameters in our approach.
Evaluation was performed using the standard {\tt \small conlleval.pl} tool for entity chunking evaluation.

In addition, we draw supervised data for multiple datasets in order to approximate the Reddit text type.
We take data from Twitter, taking the union of corpora used in previous work that follow the Freebase~\cite{bollacker2008freebase} ten-class entity scheme.
The classes given are: company, facility, geo-loc, movie, musicartist, person, product, sportsteam, tvshow and other.
For newswire, we use the Reuters RCV1 corpus annotations that were part of the CoNLL-2003 shared task~\cite{tjong2003introduction}.
Classes are removed before training, making this a \emph{chunking} task.
We evaluated performance when trained on only Twitter data; only newswire data; and also a blend of the two.
In the base cases, the same amount of data was used.
This was capped by the volume of Twitter training data available, 66k tokens; so, the newswire approach was also trained with 66k tokens.
The blended version used even amounts of both, totalling 132k data.

The baseline system was the Stanford NER tool~\cite{finkel2005incorporating}.
We included two variants: one run of the out-of-the-box stock system, using the {\tt \small english.all.3class.distsim} binary, and another with a first-order model trained on the same source data as our system.

Tuning our word representations required estimating the number of Brown clusters $C$ to use.
In prior work~\cite{derczynski2015tune}, entity recognition performance 
peaks at around $C=2500$ for corpora of 16k tokens, $C=5000$ for corpora of 32k tokens, and at higher values for larger datasets.
As this clustering is dependent on the number of types and the size of the active set $a$, and results are unreliable with $a>C$, we set $C = a = 2560$.
This trades computational cost of building clusters against the quality of the clusterings used.
We then experimented with combinations of newswire, Twitter and Reddit data.
Brown clusters are extracted using the {\tt small generalised-brown} package~\cite{sean_chester_2015_33758}.
Results are given in Table~\ref{tab:brown-tuning}.

\begin{figure}
\centering
\includegraphics[width=\columnwidth]{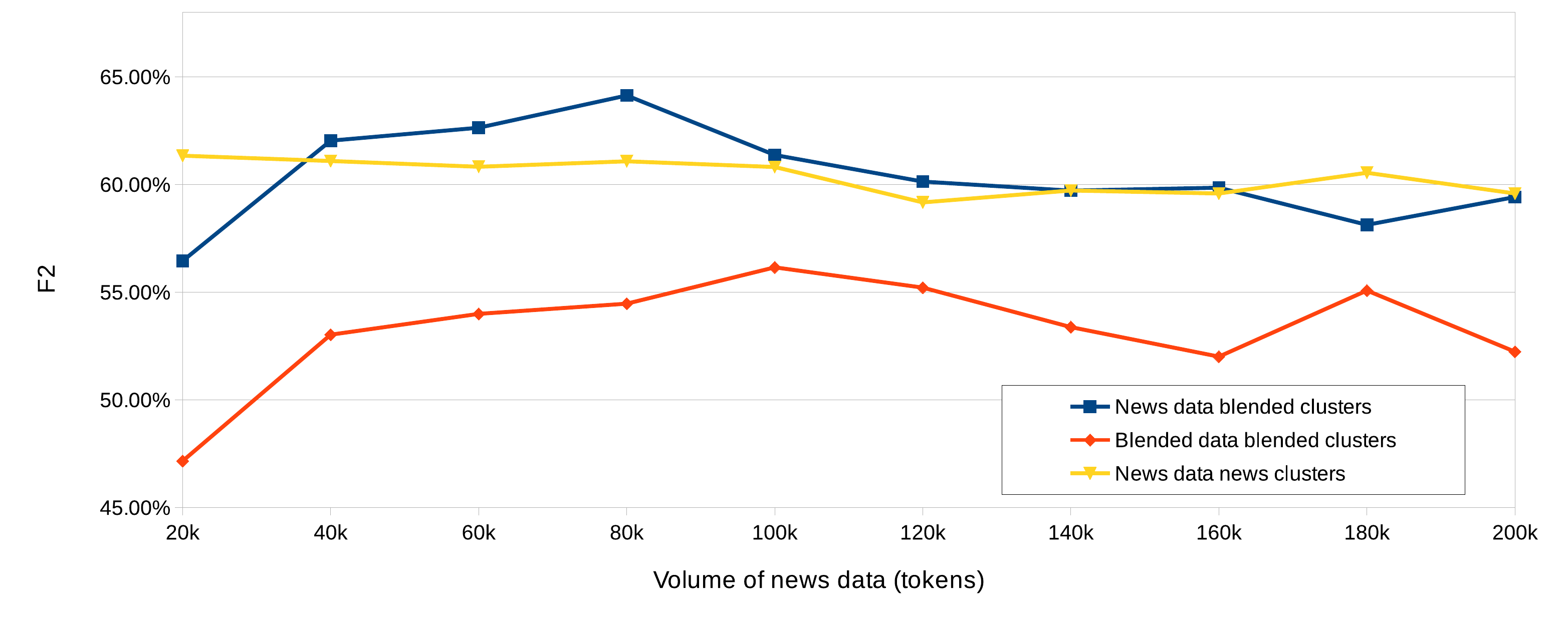}
\caption{Scaling news training data under three conditions: news clusters; blended news and Twitter clusters; blended news and Twitter clusters, with 66k extra Twitter training data. Note that increasing the proportion of news training data led to decreased performance in every case.}
\label{fig:rcv-scaling}
\end{figure}

Note how the scores are consistently best in each category when inducing Brown clusters from Reddit data.
Attempts to approximate this using newswire, Twitter, or a blend of those two did not score as well.
This is remarkable considering that we used only 64M tokens of Reddit data for cluster induction, compared to around 130M total for the other text type blends.
So, half the amount of in-text-type data provides notably improved unsupervised representations over approximated supervised data.

We experimented with pure newswire and also newswire plus tweet training data, and with pure newswire vs. blended clusters.
Results are given in Figure~\ref{fig:rcv-scaling}.
Blending training text types in supervised learning did not lead to improved performance.
This suggests that adding too much newswire reduces performance, and that limiting to just newswire clusters also reduces performance.
Based on this data, we hypothesised that insufficient regularisation had led to overfitting.
To test this, noting the downward turn in newswire-trained blended cluster performance after 80k tokens (Figure~\ref{fig:rcv-scaling}), we re-ran the experiments with 100k newswire data using a $c_2$ regularisation penalty of $10^{-1}$. 
Performance rose to F2 62.07\%; so, while effective, still a marginal increase.
Therefore, we continued using newswire training data with Reddit clusters for annotating the named entities across the 100 sampled subreddits.

\section{Entity Diffusion}
\label{sec:diffusion}
In this section we now move on to examining how the recognised named entities emerge and \emph{diffuse} through the analysed subreddits.
As per prior work, one of the first things that we can inspect is the \emph{shape} of entity mention cascades: that is, the patterns of diffusion that such entities exhibit when cited in conversation chains. 
We begin by explaining how such patterns are derived, before then moving on to showing what patterns emerge.

\subsection{Entity Mention Cascades}
Prior work by Leskovec et al.~\cite{leskovec2007patterns} examined the shapes of hyperlink cascades through the blogosphere to identify patterns of link diffusion.
We follow a similar process here, however we instead inspect the emergence of entities in conversation chains in Reddit.
We first make the following explicit.

\begin{mydef}
(Entity Cascade) A cascade of ${<p_i, pj>} \in C_e$ of an entity $e \in E$ occurs when two or more posts citing the entity are chained together in a reply graph. 
Hence: $C_e = \{<p_i, p_j> : p_i \rightarrow p_j \in R, cites(p_1) = cites(p_j) = e\}$.
\end{mydef}

Our goal is to derive all cascades for each entity in our analysis, and then examine how the shapes and sizes of these cascades differ.
To gather each entity's cascades, we retrieved all (of the 100) subreddit posts that contained a given entity $e$.
Then, for each post ($p \in P_e$) we recovered the reply-chain that that entity appeared within -- this was performed by going \emph{up} the reply chain from $p$ to its parent post (i.e. the post that $p$ was replying to) and \emph{down} the reply chain by getting the posts that replied to $p$.
When iterating through the posts, if we came across a post that replied to another post in an existing chain then that post was added to the chain.
We only maintained posts within the chain that cited the entity in question: this produced entity cascades where each consecutive post in the chain mentions the entity -- we refer to this as \emph{strict cascade derivation}, as we do not consider posts higher-up or lower-down the reply chain that cite the entity yet are connected by a non-entity citing post.\footnote{Chain-derivation Python code can be found here: \url{https://github.com/mrowebot/NER-Diff-Paper}}

\begin{figure}[ht!]
  \begin{center}
  \subfigure[Top-20 Cascade Shapes]{\includegraphics[width=0.49\columnwidth]{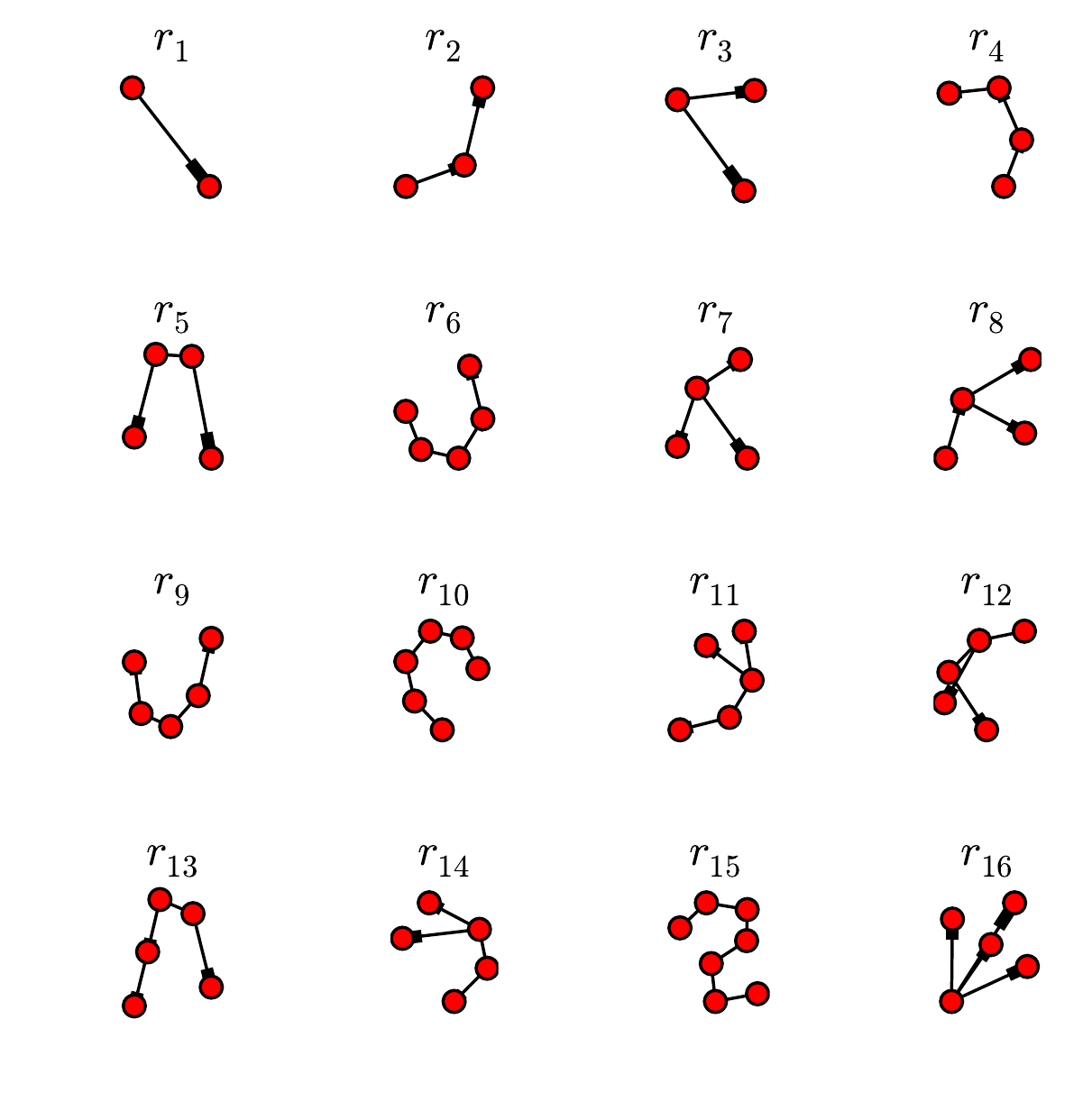}\label{fig:cascade_shapes}}   
  \subfigure[Cascade Shape Rank Distribution]{\includegraphics[width=0.49\columnwidth]{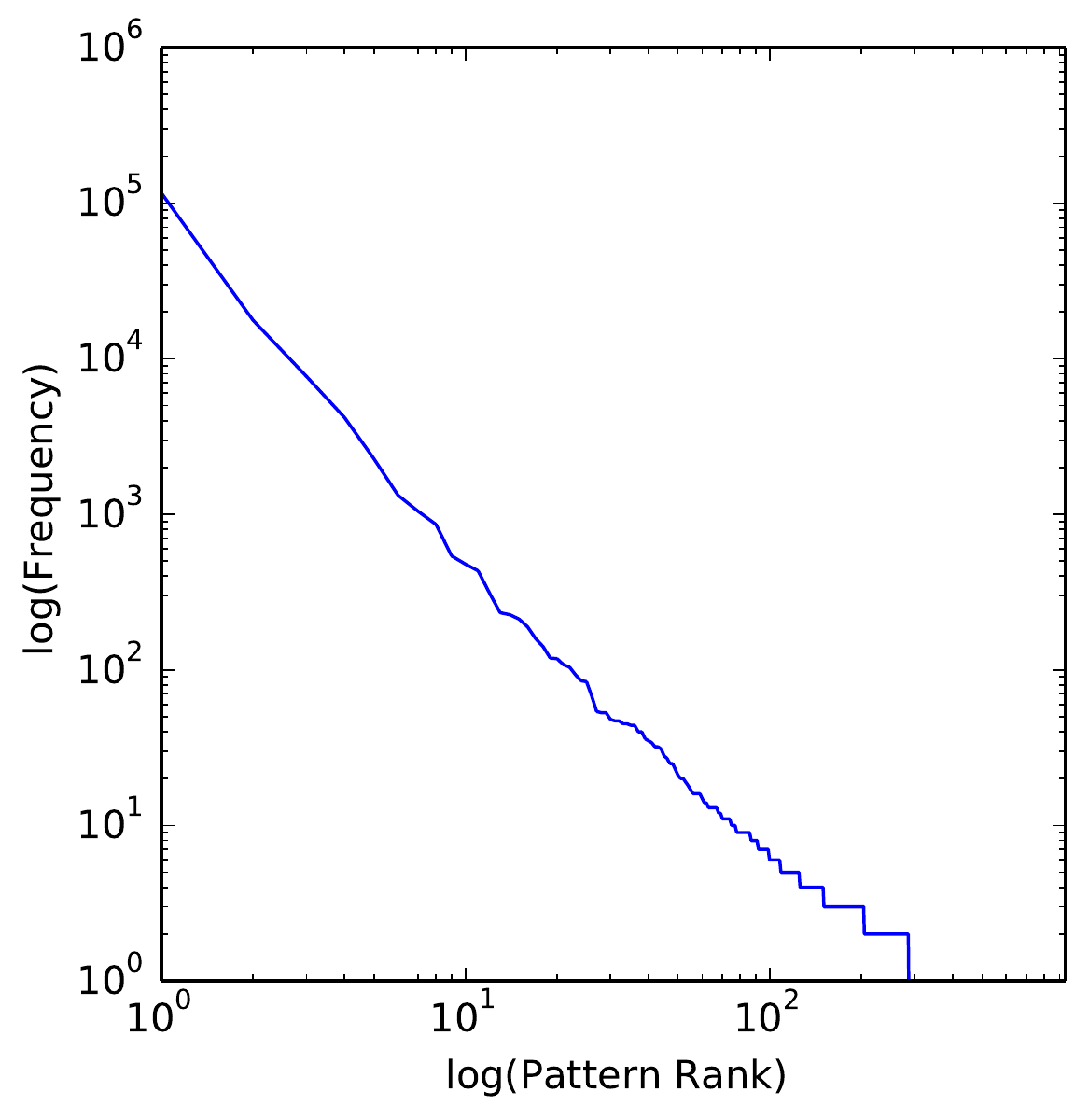}\label{fig:rank_dist}}
  \end{center}    
  \caption{The top-20 cascade shapes are generally deep and narrow with little branching (Fig.~\ref{fig:cascade_shapes}, while the cascade shape rank follows a power-law distribution (Fig.~\ref{fig:rank_dist}.}
  \label{fig:entity_cascades}
\end{figure}

This process produces, in essence, a collection of cascade graphs for each entity, each of which may have isomorphic shapes yet contain different node labels (i.e. different post ids).
We reduced each entity's cascade graph collection down to a frequency distribution of the \emph{canonical form} of each graph using Cordella at al.'s~\cite{cordella2001improved} graph isomorphism approach.
A further reduction was run to compile a frequency distribution of the cascade shapes across all entities.
Fig.~\ref{fig:entity_cascades} show both the top-20 entity cascade shapes on the left (Fig.~\ref{fig:cascade_shapes}) and the ranking of the patterns' frequencies on a log-log scale (Fig.~\ref{fig:rank_dist}).
Upon inspection, one thing becomes immediately apparent: entity cascades are shallow and short at the top-3 ranks, however after this position we start to see chains of discussions as being popular which are deeper and narrower.
This result contrasts prior work~\cite{leskovec2007patterns} where cascades of hyperlinks between blogs were shallower in depth yet wider -- in terms of the breadth of diffusion at the first level from the seed. 
The ranking of the patterns follows a general power-law distribution where a small section of patterns (i.e. the top-20) are seen most often -- this is somewhat expected as it would be very rare for an entity to be cited in a long thread with many branching reply-chains.

\subsection{Entity Adoption Post-$({k-1})^{th}$ Exposure}
Inspection of the shape of entity cascades through Reddit discussion threads reveals some interesting traits, suggesting that an entity \emph{spreads} through narrow diffusion paths -- i.e. with little branching occurring.
One natural question that emerges from this is to question the extent to which exposures to an entity play a role in actually adopting (i.e. citing) the entity in question.
To investigate the relationship between exposures and adoptions, we took the top-500 entities from our whole annotated dataset and calculated the probability of a user adopting an entity after being \emph{exposed} to the entity $k$ times, defining an exposure as follows:

\begin{mydef}
\label{def:exposure}
(Exposure) A user $u$ is exposed to an entity $e$ at time $t$ if a given post $p \in P^{\Gamma(u)}$ authored by a neighbour of $u$ (i.e. $v \in \Gamma(u)$) contains the entity $e$, where neighbours interacted with $u$ prior to $t$.
\end{mydef}

\begin{figure}[ht!]
  \begin{center}
  \subfigure[Exposure-Adoption Global Distribution]{\includegraphics[width=0.49\columnwidth]{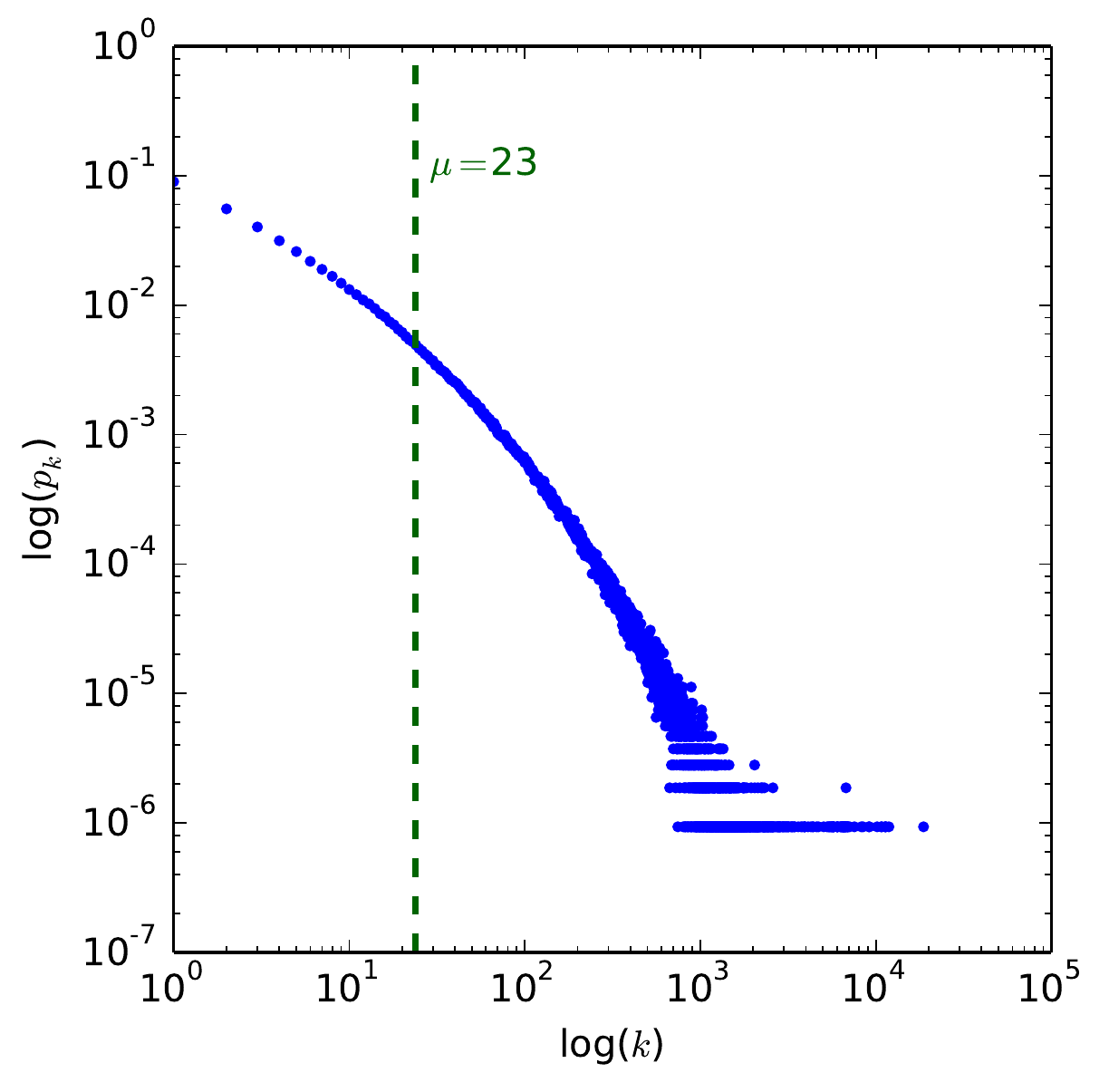}\label{fig:global_exposure_dist}}   
  \subfigure[Sampled Entity Exposure-Adoption Distributions]{\includegraphics[width=0.49\columnwidth]{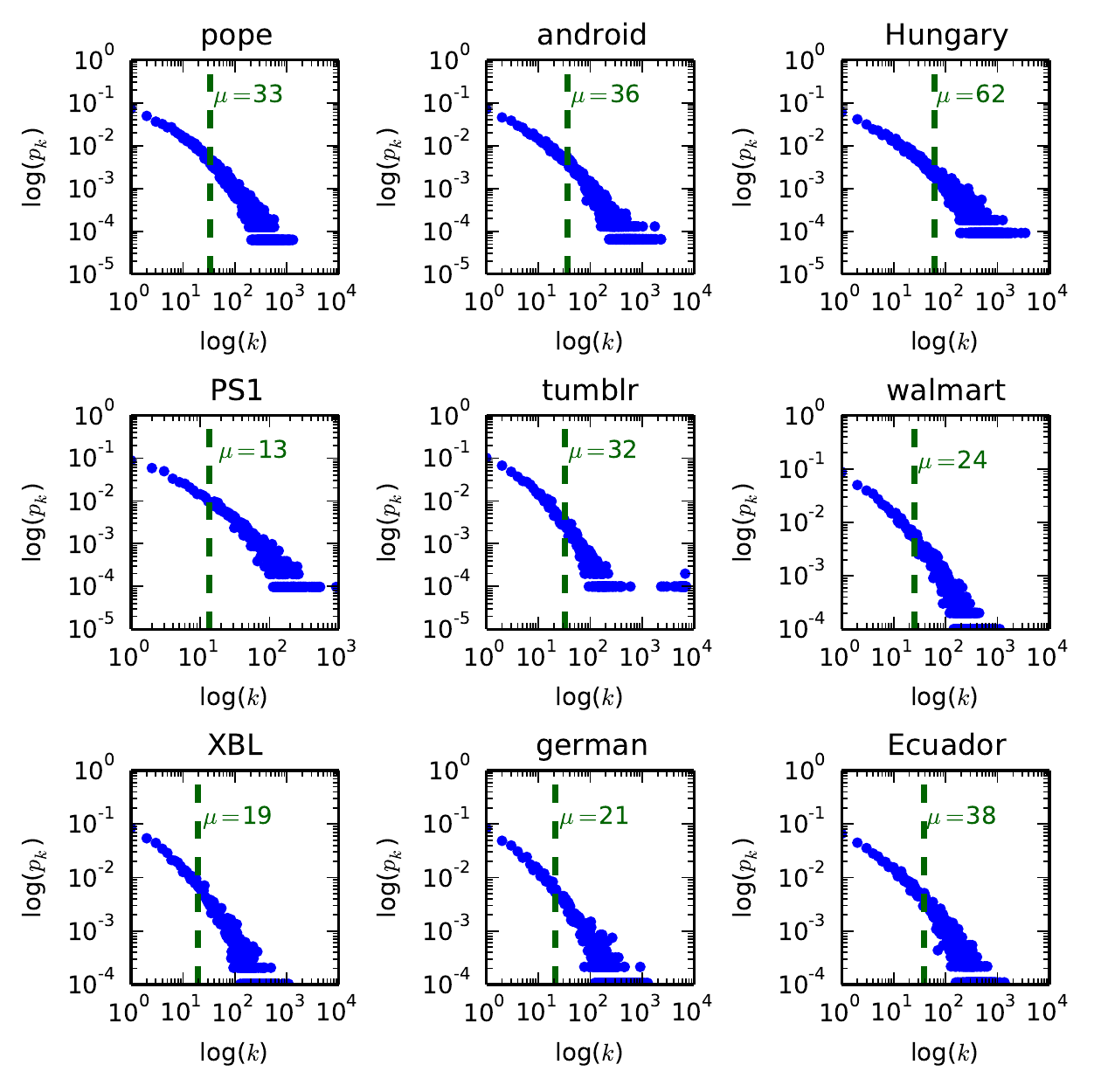}\label{fig:entity_exposure_dists}}
  \end{center}    
  \caption{The probability of a user adopting an entity as a function of $k$ prior exposures to the entity has a heavy-tailed distribution (Fig.~\ref{fig:global_exposure_dist}) that is consistent across all entities, including a sample of 9 random entities (Fig.~\ref{fig:entity_exposure_dists}).}
  \label{fig:exposure_dists}
\end{figure}

Based on this definition we iterated chronologically through all posts that cited a given entity.
If the post was the first time that a user cited the entity (i.e. he/she was not \emph{activated}) then we counted how many \emph{exposures} the user had received prior to the time of the post -- logging this as $k$.
Fig.~\ref{fig:global_exposure_dist} presents the overall plot of the probability (i.e. relative frequency) of users adopting an entity after $k$ exposures to the entity.
Immediately, one can note that the mode of this distribution is at $0$ and that the mean is $k=23$: this implies that users are most likely to actually cite an entity without having been exposed to it, in fact $P(adoption) \rightarrow 0, k \rightarrow \infty$.
We are somewhat guarded in \emph{generalising} from this result, as our experimental setup here -- given the scale of the data we are playing with and the tractability of annotating the entirety of Reddit -- results in only a fraction of Reddit being annotated with entities.
Hence, it is possible that entities emerge from other subreddits, yet we are unable to capture this at present -- our future work suggests how this effect can be validated.
Furthermore, this finding contrasts somewhat to existing patterns of \emph{hashtag} adoption~\cite{romero2011differences} where there is a clear mode at around $k=4$ exposures, after which the probability of adoption curtails.
This difference is likely due to two factors.
Firstly, the difference between platforms; Twitter acts as public broadcast where information is presented in feeds and is then passed on, while Reddit is more interaction and discussion-driven.
Secondly, the manner in which users are \emph{exposed} to information; on Twitter this is via subscriptions to other users and observing trends in the trending topics area, while Reddit requires users to read through threaded discussions and \emph{notice} entities within.

The second plot below (Fig.~\ref{fig:entity_exposure_dists}) shows a sample of 9 entities' adoption-exposure distributions, all of which have similar shapes (with a mode at $0$) and a heavy tail.
There is variance in the means of these distributions.
For instance, the entity \emph{PS1}\footnote{Denoting the original Playstation video-games console.} has a lower mean than the entity \emph{Hungary}, suggesting  users require less stimulation to discuss the former than the latter.
The nature of how and why the distributions differ is something that requires further investigation.

\subsection{Global Threshold Diffusion Model}
We now move to forecasting the diffusion of entities across Reddit.
For this, we used a modified implementation of Goyal et al.'s general threshold model~\cite{goyal2010learning} to parallelise computation of the model.
The core principle of the model is that one can calculate the probability of a user ($u$) adopting an entity ($e$) based on how their neighbours ($v \in \Gamma(u)$) have influenced them previously.
Hence, the probability of $u$ adopting an entity is calculated as follows:

\begin{equation}
\label{eq:joint_prob}
p_u\big(\Gamma(u)\big) = 1 -- \displaystyle\prod_{v \in \Gamma(u)}\big(1 -- p_{v,u}\big)
\end{equation}

In Goyal et al.'s prior framework, the probability of influence ($p_{v,u}$) of $v$ on $u$ is based upon the maximum likelihood estimate of a single Bernoulli trial.
An entity propagation occurs from $v$ to $u$ when the latter cites $e$ after being exposed to it by the former (as per Definition~\ref{def:exposure}), hence a count of how many entities propagate between $v$ and $u$ can be recorded in $E_{v2u}$.
So, the influence probability between $v$ and $u$ based on such \emph{propagation} can be calculated thus, where $E_v$ is how many times $v$ has cited an entity:

\begin{equation}
\label{eq:action_prob}
p^{E}_{v,u} = \frac{E_{v2u}}{E_v}
\end{equation}

The authors present two variants of this calculation: (i) a static Bernoulli random trial where Equation~\ref{eq:action_prob} is calculated from the training set, ignoring time; (ii) a discrete time Bernoulli random trial where counts are only placed within $E_{v2u}$ and $E_v$ if the citation of an entity is within a discrete time interval, that is: if the time that $u$ adopts an entity $e$ is given by time $t_u$ then $E_{v2u}$ and $E_v$ are composed from the entity posts of $v$ which each have time $t_v \in [t_u -- \tau_{v,u}, t_u)$, where $\tau_{v,u}$ is derived as follows (only considering $v, u \in U$ (set of all users) if $u$ has contacted $v$ prior to $t_u$:

\begin{equation}
\tau_{v,u} = \frac{\displaystyle\sum_{e \in E} (t_u(e) -- t_v(e)}{E_{v2u}}
\end{equation}

\begin{figure}[ht!]
  \begin{center}
  \subfigure[Binned Frequency Distribution of $\tau_{v,u}$ values]{\includegraphics[width=0.49\columnwidth]{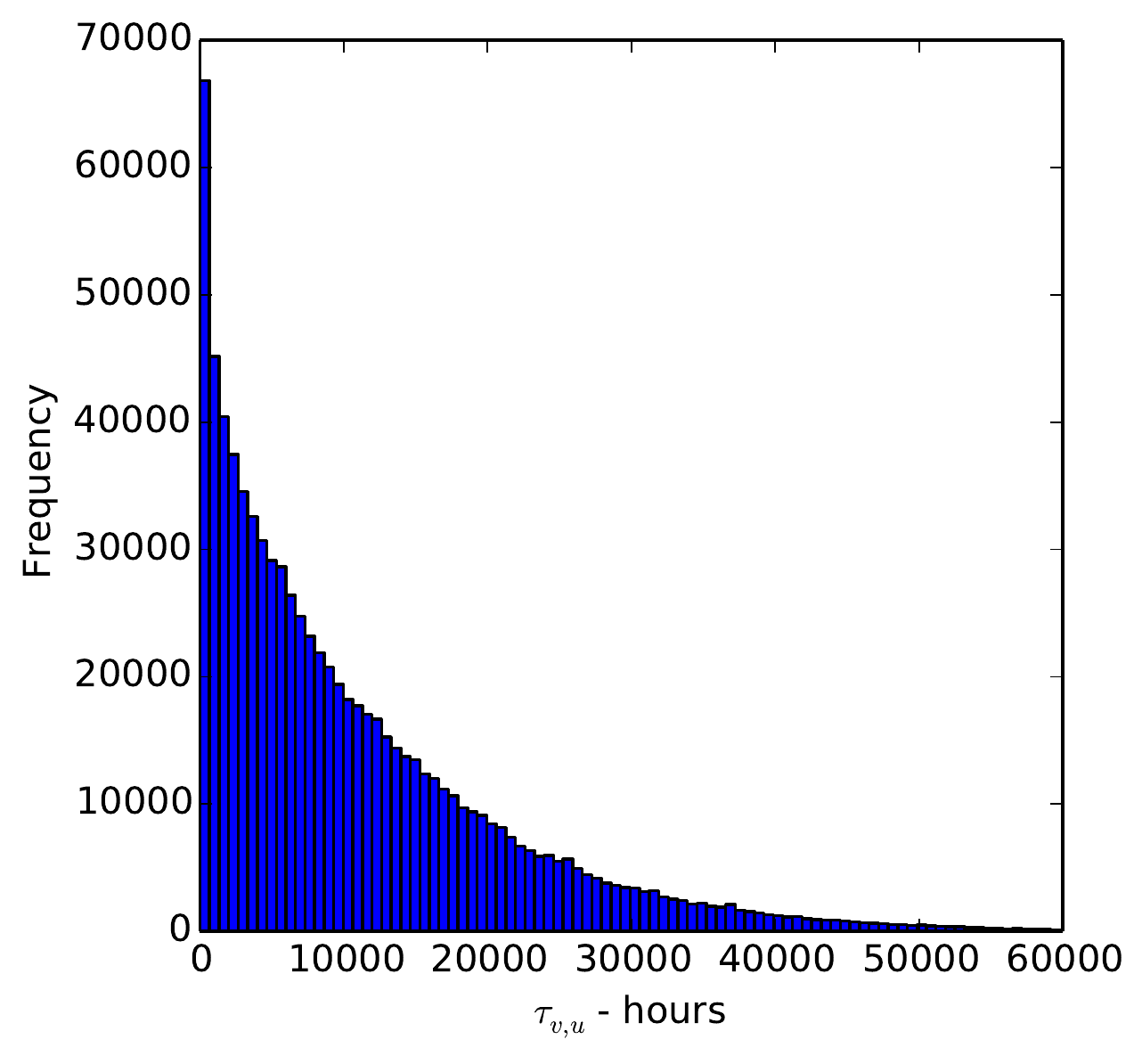}\label{fig:taus_hist}}   
  \subfigure[Log-log Plot of the Binned $\tau_{v,u}$ distribution]{\includegraphics[width=0.465\columnwidth]{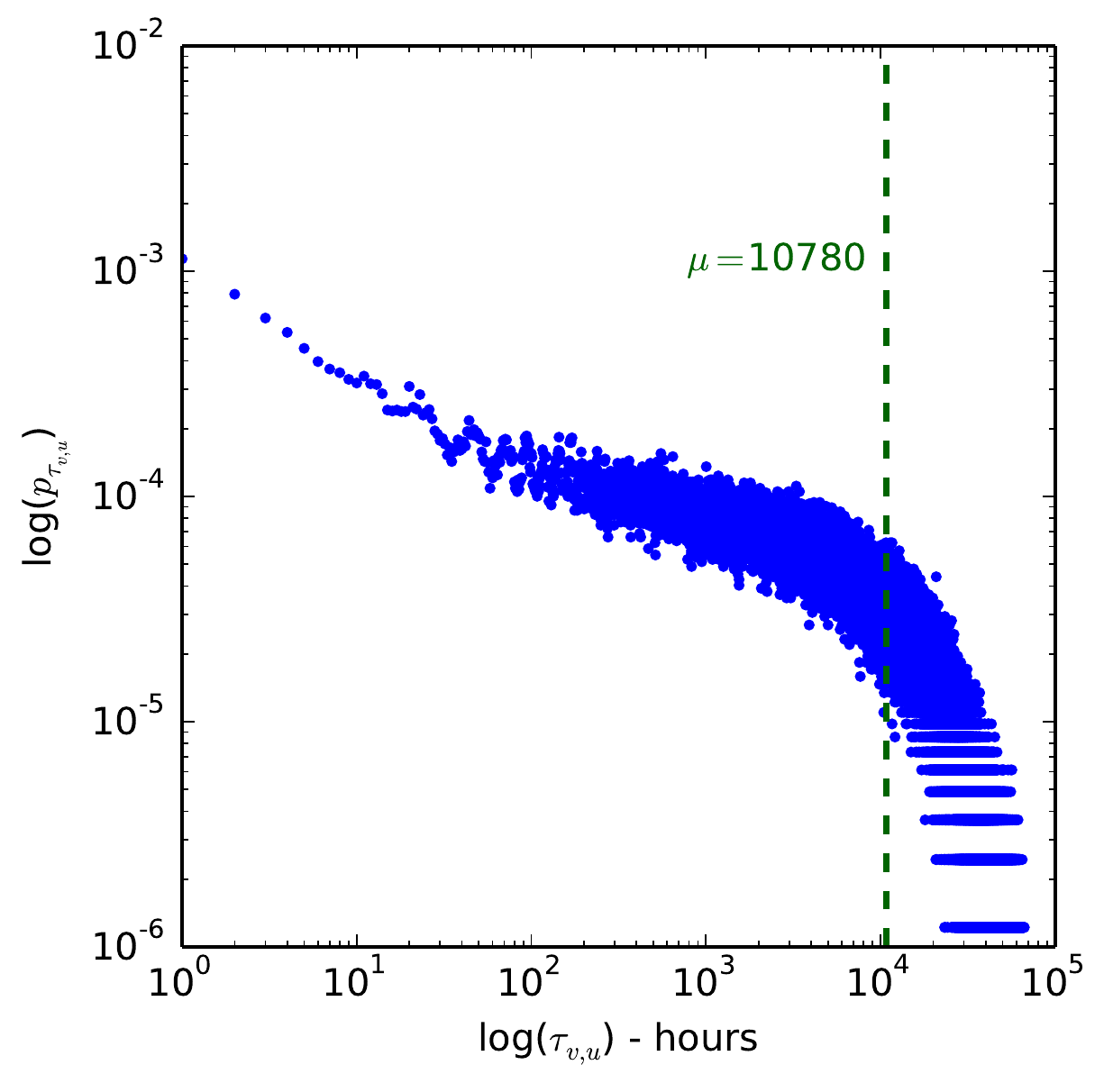}\label{fig:taus_loglog}}
  \end{center}    
  \caption{The influence window ($\tau_{v,u}$) between two arbitrary users characterises the average time for an entity to propagate from $v$ to $u$. In hours, this value has a \emph{right-skew} (Fig.~\ref{fig:taus_hist}), while the log-log plot (Fig.~\ref{fig:taus_loglog}) of the relative frequency distribution demonstrates the heavy-tail nature of the distribution with a mean of $10,780$ hours ($\approx 449$ days $\approx 1.2$ years).}
  \label{fig:taus}
\end{figure}

Fig.~\ref{fig:taus_hist} shows the binned distribution of the $\tau_{v,u}$ values. Note the distribution has a right skew with the mode of the distribution being around one hour; this then gradually tails off with fewer people having larger \emph{influence windows}.
The log-log plot of the relative frequency distribution (Fig.~\ref{fig:taus_loglog}) shows the \emph{heavy-tail} property of the distribution, and that the mean window width is $10 780$ hours ($\approx 449$ days $\approx 1.2$ years), indicating a degree of \emph{stickiness} in Reddit communities, where people remain for long periods.

\subsubsection{Additional Influence Dynamics -- Entity-Adoption Constructs}
The neat formulation of the general threshold model, and the monotonic-submodular nature of the probability of adoption function ($p_u(\Gamma(u)$), means that we can vary the mechanism by which we derive the \emph{influence probability} ($p_{v,u}$) between two users $v$ and $u$ to test for different influence effects -- we refer to these as \emph{entity-adoption constructs}.
Our contribution here is to test for the influence of \emph{prior interactions} and \emph{community homophily} using the general threshold model.
To compute the influence probability based on interactions, we derive $p^I_{v,u}$ as follows:

\begin{equation}
p^{I}_{v,u} = \frac{\vert \{ p_u : p_v \in P_v, p_u \in P_u, p_u \rightarrow p_v \} \vert}{\vert \{ p_u : p_u \in P_u, p_u \rightarrow . \} \vert}
\end{equation}

Where $P_u$ and $P_v$ denote the set of posts by users $u$ and $v$ respectively, and $p_u \rightarrow p_v$ indicates that post $p_u$ replied to post $p_v$.
The influence probability based upon community homophily ($p_{v,u}^C$) is derived as follows:

\begin{equation}
p^C_{v,u} = \frac{\vert C_u \cap C_v \vert}{\vert C_u \cup C_v \vert}
\end{equation}

Where $C_u$ and $C_v$ are the sets of subreddits that $u$ and $v$ has posted in respectively.
We calculate the same two variants as the entity-propagation influence probability as above (static Bernoulli, and discrete-time Bernoulli), for the discrete-time case we only consider interactions between $u$ and $v$ that fall within $[t_u -- \tau_{v,u}, t_u)$ (interactions-based) and posts within subreddits by $u$ and $v$ that were made within $[t_u -- \tau_{v,u}, t_u)$ (community-homophily).

In order to eliminate bias in our below experiments -- when we attempt to forecast entity adoption for users -- we divided the top-500 entities into an 80\%:20\% split for training and testing respectively.
Then, for the above influence probabilities (entity-propagations, interactions-based, community-homophily) we used different strategies for their calculation.
For the entity-propagation influence probability ($p^E_u(\Gamma(u))$) we used the training segment to calculate the values of $E_{v2u}$ and $E_v$, and also $\tau_{v,u}$ -- for all pairs of interacting users within the training segment -- this follows the experimental setup of~\cite{goyal2010learning}.
One thing that is somewhat limited about this approach, is that we are observing future effects when calculating $E_{v2u}$ and $E_v$ that we take forward into our experiments, as we observe how influence has occurred between users prior to an adoption happening.
This is somewhat unavoidable in the context of dataset splitting as $\tau_{v,u}$ must be calculated somehow -- an alternative for future work is to use a fixed time-split and use the first 80\% of entity-posts for training and the rest for testing.

\subsection{Experiments}
We now move on to forecasting the adoption of named entities by users as they spread through Reddit.
To this end, we used an experimental setup that induces the joint probability function (Eq.~\ref{eq:joint_prob}) on a per-entity basis within the test set: each user's probability of adoption was computed as product of their neighbours' influence.
Our goal therefore was to examine which of the above entity-adoption constructs were best suited to predicting adoptions.

\subsubsection{Experimental Setup}
Using the 100 randomly sampled subreddits and running the above Named Entity Recogniser over these subreddits' posts resulted in over $300$ million posts in our dataset (using only those from the 100 subreddits) written by $4 139 814$ users -- the entity recogniser also extracted $8 797 271$ unique entities.
For our experiments we tested 6 different models that resulted from permutations of the 2 probability settings (i.e. static Bernoulli or discrete-time Bernoulli (i.e. $t_u \in [t_u -- \tau_{v,u}, t_u)$) and the 3 entity-adoption constructs (entity-propagation, interactions-based, and community-homophily).

\paragraph{Deriving Adoption Probabilities}
In order to test which model was best (from above) we took the entities within the test set, and ran the following process: we chronologically ordered each entities' posts and then iterated through the posts set one-by-one.
For each post's author ($v$) we then checked if they had been \emph{activated} before -- i.e. had they cited the entity? -- if not, then this would be first time they had cited $e$.
If this was the case then we retrieved the prior interactions that the user had had and calculated (for each prior neighbour -- $u \in \Gamma(v, t_u)$) the probability of influence between $v$ and $u$ using the above influence probability variants (e.g. interactions-based with static Bernoulli setting).
We then updated the probability of adoption of $u$.
By iterating through the set of time-ordered posts we maintained adoption-outcome tuples of the form $<u, p_u, r_u>$ where $r_u \in \{0,1\}$ denoting whether the user ultimately adopted the entity $e$ or not.
Our evaluation of the models used these tuples to calculate the area under the Receiver Operator Characteristic ($ROC$) curve, aiming to achieve a value of $1$ (for perfect prediction).

\paragraph{Parallelising Processing}
As we are working with large datasets (i.e. $>300$ million posts), we made two efforts to parallelise induction of adoption probabilities over test set entities.
First, of the data used (timestamped interactions between users, entity posts, post details, $E_{v2u}$ values, $\tau_{v2u}$ values) was uploaded into HBase tables. 
Second, we used Apache Spark
 to parallelise the per-entity diffusion processes.
This involved loading the names of the test entities into HDFS and then forcing Spark to partition the entity list into at least 30 partitions.
Each partition was then iterated over and the above test process run: (i) retrieving time-ordered entity posts from HBase, (ii) iterating over the post set, (iii) retrieving per-user interactions prior to the time of a given post, and (iv) calculating the pairwise influence probabilities.
The final calculated probability of adoption for each user ($u$) and the label of whether they adopted the entity or not were recorded in a separate HBase table.

Due to the use of a sample of 100 of the top-500 entities in our experiments, iteration over the time-ordered post set required an expensive sequential scan -- which cannot be avoided.
That said, we were able to add a second level of parallelism however, given the sub modular and monotonic nature of the joint probability as follows.
Calculation of the probability of $e$ being adopted by $u$ is derived from Eq.~\ref{eq:joint_prob}, and is calculated from the prior neighbours of $u$ before adoption.
Now, as this function is sub-modular and monotonic, we could \emph{update} the probability of adoption given a new neighbour's ($v$) influence probability as follows:

\begin{equation}
p_s(\Gamma(u) \cup p_{v,u}) = p_s(\Gamma(u)) + (1 -- p_s(\Gamma(u))) * p_{v,u}
\end{equation}

Also, \emph{multithreading} the calculation of the influence probabilities between $v$ and each of their neighbours $u \in \Gamma(v)$ gave additional parallelism. We calculated these pairwise influence probabilities in parallel and then updated $p_u(\Gamma(u)) : \forall u \in \Gamma(v)$.\footnote{Maintenance of interactions between users stores both interaction source and target. Thus we can retrieve directed interactions both ways -- i.e. $v \rightarrow u \wedge v \leftarrow u$.}
Java implementation of this code can be found in the github repository,\footnote{\url{https://github.com/mrowebot/NER-Diff-Paper}} including the functions for building the HBase tables, deriving the entity-propagation counts ($E_{v2u}$) and the test algorithm.

\begin{table*}
\begin{center}
\caption{Area under the Receiver Operator Characteristic Curve (ROC) values for the different probability settings and influence probability settings within the general threshold model.}
\begin{tabular}{ l | c c | c c}
\hline
& \multicolumn{4}{c}{Probability Setting} \\
Entity-adoption Construct & \multicolumn{2}{c |}{Static} & \multicolumn{2}{c}{Discrete-Time} \\
 & Micro-$ROC$ & Macro-$ROC$ & Micro-$ROC$ & Macro-$ROC$ \\
\hline
Entity-propagations ($p_{u}^{E}$)  & $0.730$ & $0.713 (\pm0.095)$ & $0.730$ & $0.714 (\pm0.096)$ \\
Interactions-based ($p_{u}^{I}$) & $0.755$ & $0.710 (\pm0.095)$ & $0.666$ & $0.644 (\pm0.091)$ \\
Community-homophily ($p_{u}^{C}$) & $0.715$ & $0.740 (\pm0.147)$ & $0.643$ & $0.631 (\pm0.085)$  \\
\hline
\end{tabular}
\label{tab:macro_results}
\end{center}
\end{table*}

\subsubsection{Results}
Table~\ref{tab:macro_results} presents the results from our experiments of the various entity-adoption constructs and probability settings, including the micro- and macro-averaged $ROC$ values and deviations for each model.
The micro-averages are computed by \emph{pooling} together all result tuples (i.e. $<u, p_u, r_u>$ ) from all the test entities, and working out the $ROC$ value from that pool. The macro-averages are computed by working out the entity-specific $ROC$ values and deriving the mean (and standard deviation) from those.

Overall the results indicate that static Bernoulli probability achieves the best performance. The best performing of the models are the Interactions-based (from Micro-$ROC$) and the Community-homophily (from Macro-$ROC$).
We note the following:

\begin{itemize}
	\item The window of influence (characterised by $\tau_{v,u}$) is too narrow, as emphasised by Fig.~\ref{fig:taus_hist}. 
	While static probabilities capture influence over a large time-period, they actually contribute information about influence between users based on interactions and the similarity of communities they post within. 
	As a result, the window omits interactions prior to this.
	The performance difference between interactions-based and community-homophily models reflects this property.
		
	\item Interactions have the greatest effect on adoption, not prior entity adoptions.
	The static Bernoulli model indicates that, in general, adoption of an entity is influenced by the intensity of interactions between two individuals, and not necessarily just whether propagation has actually occurred.
	This finding reflects the communal nature of Reddit, where users constantly follow-up to posts with comments, which then evolve into a threaded discussion.
	It is likely that, in this context, interactions around specific topics (within designated subreddits) occur frequently between clusters of users, thereby leading towards discussions around certain entities later.
	
	\item Adoption from community homophily varies between entities.
	The (relatively) large standard deviation for the community homophily model with static Bernoulli setting in Table~\ref{tab:macro_results} indicates how varied community-homophily can be.
	One could hypothesise here that entities which are specific to a given community and/or are emergent within a community would require a user to be \emph{similar} to their peers in order to adopt it from them; whereas general entities are more likely to be ignored.
	
\end{itemize}

\section{Discussion and Future Work}
\label{sec:discussions}
This paper presents one of the first pieces of work examining how entities spread through social networks.
As a result of this novelty, our work has prompted a variety of avenues for future work.
Therefore in this section we reflect on the approach we adopted and any potential issues that may arise from this, before then outlining our future work plans.
One of the core findings that we presented in this work is that the mode of the exposure-adoption function (i.e. probability of adoption as a function of $k$ exposures to an entity) resides at $k=0$.
As we had to restrict the annotation of Reddit to a sample of 100 subreddits, it is possible that users were \emph{exposed} to entities beforehand but within communities that we did not annotate.
Therefore to validate our finding, extended work will take a sample of $1,000$ entities that match those entities within the whole of the Reddit dataset.
The exposure-adoption graphs will then be derived once again from this information.

Our second proposal for future work is to extend the univariate deterministic case that we have explored thus far -- i.e. calculating the probability of $u$ adopting $e$ -- to a multivariate case -- i.e. calculating the probabilities of $u$ adopting members from entity set $E$, where members of $E$ are \emph{colinear}.
This allows for the investigation of \emph{vulnerability} windows to be explored, which would characterise how \emph{susceptible} a given user is to adopting an entity (or any colinear contagion) based on their recent adoptions.
The third future work effort will be to extend  calculation of adoption probabilities to the continuous time case -- as in~\cite{huang2014temporal} -- by computing the probability of one user \emph{influencing} another user based on the latency to their latest interaction.
This allows entity adoption to be explored from the perspective of pairwise interactions, as opposed to entity-propagations -- potentially alleviating the confounding effect of the influence window we found in the discrete-time setting.

\section{Conclusions}
\label{sec:conclusions}
Understanding how entities spread through social networks provides researchers and marketers with valuable insights to recover and forecast the diffusion process.
Our study of entity diffusion began by presenting an accurate means to obtain named entities from within discussion posts, before moving on to examining what \emph{patterns} of entity diffusion occur -- and how frequently -- and how exposures to entities are associated with the probability of a user adopting an entity.
Following these findings we presented a general threshold diffusion model that allows for different entity-adoption constructs to be tested within the diffusion process: our results from applying this model indicated that the interactions between individuals provide the most accurate means of calculating influence probabilities and thus forecasting entity adoption.

In the introduction of this paper we set forth three research questions. 
We now revisit these questions and highlight the evidence presented in our paper and how this has contributed towards answering the questions:

\textbf{RQ1:} How can we accurately detect named entities in social media based discourse, given its myriad formats, often informal vernacular, and inherent noise (e.g. misspellings, abbreviations, etc.)?
We have presented a method to detect named entities within Reddit posts that uses structured prediction and Brown clustering.
Furthermore, we presented an empirical evaluation of our method when trained using a blend of named entity annotated corpora to transfer existing annotations from disparate corpora (covering different language styles) as training data.
We found that representations induced from in-genre unsupervised data were much more helpful than approximating the supervised data by mixing other genres.

 \textbf{RQ2:} What process governs the spread of entities? And how does such spread occur?
We derived key insights into what diffusion patterns are found when entities spread through threaded discussions,  finding that, unlike the spread of hyperlinks in the blogosphere~\cite{leskovec2007patterns}, entities exhibit relatively \emph{deep} and \emph{narrow} diffusion traces.
We also investigated the association between the number of exposures that users receive of an entity and the probability of a user adopting said entity thereafter, discovering that adoption probability decays as exposure count increases.

 \textbf{RQ3:} How can we predict the spread of named entities and who will begin talking about them?	
Putting all the pieces together, we implemented a modified version of a general threshold model which incorporated entity-adoption constructs to test different mechanisms for computing user-to-user influence probabilities and can be learnt in parallel.
Our empirical evaluation of this framework found that interactions had the greatest \emph{overall} effect, while there was variance between entities in terms of the impact of \emph{community-homophily} on users adopting an entity.

\bibliographystyle{ACM-Reference-Format}
\bibliography{NER-Diffusion}

\end{document}